# MD-Syn: Synergistic drug combination prediction based on the multidimensional feature fusion method and attention mechanisms

XinXin Ge[1], Yi-Ting Lee[2], and Shan-Ju Yeh[1,2,3,*]

[1] School of Medicine, National Tsing Hua University, Hsinchu, Taiwan.
[2] Institute of Bioinformatics and Structural Biology, National Tsing Hua University, Hsinchu, Taiwan.
[3] Department of Life Science, National Tsing Hua University, Hsinchu, Taiwan.

[*] Corresponding author.
E-mail: sjyeh@life.nthu.edu.tw (Shan-Ju Yeh)

**Abstract**

Drug combination therapies have shown promising therapeutic efficacy in complex diseases and have demonstrated the potential to reduce drug resistance. However, the huge number of possible drug combinations makes it difficult to screen them all in traditional experiments. In this study, we proposed MD-Syn, a computational framework, which is based on the multidimensional feature fusion method and multi-head attention mechanisms. Given drug pair-cell line triplets, MD-Syn considers one-dimensional and two-dimensional feature spaces simultaneously. It consists of a one-dimensional feature embedding module (1D-FEM), a two-dimensional feature embedding module (2D-FEM), and a deep neural network-based classifier for synergistic drug combination prediction. MD-Syn achieved the AUROC of 0.919 in 5-fold cross-validation, outperforming the state-of-the-art methods. Further, MD-Syn showed comparable results over two independent datasets. In addition, the multi-head attention mechanisms not only learn embeddings from different feature aspects but also focus on essential interactive feature elements, improving the interpretability of MD-Syn. In summary, MD-Syn is an interpretable framework to prioritize synergistic drug combination pairs with chemicals and cancer cell line gene expression profiles. To facilitate broader community access to this model, we have developed a web portal (https://labyeh104-2.life.nthu.edu.tw/) that enables customized predictions of drug combination synergy effects based on user-specified compounds.

**Key points**
- We proposed a novel computational framework, MD-Syn, which achieved the

- AUROC of 0.919 in five-fold cross-validation for conducting synergistic drug combination prediction based on multidimensional feature fusion methods and multi-head attention mechanisms.
- For one-dimensional feature embedding module (1D-FEM) in MD-Syn, we applied a large-scale chemical language pre-trained model and multi-layer perceptron classifier to have representations for small molecules and cancer cell lines, respectively.
- For the two-dimensional feature embedding module (2D-FEM) in MD-Syn, we used a graph convolutional network (GCN) to obtain the graph representation for each drug, and leveraged the node2vec algorithm to learn the node embedding for each protein in the protein-protein interaction (PPI) network.
- We designed a trans-pooling block in 2D-FEM with multi-head attention mechanisms that could not only capture representations from different feature aspects but also improve the interpretability of MD-Syn.
- Integrating two modalities of each input feature type, the MD-Syn improved 2.6% in AUROC on synergistic drug combination prediction.

## INTRODUCTION

Most human diseases are caused by complex biological processes that cannot be cured entirely by a single drug treatment strategy. Compared to single-agent therapies, drug combinations have the potential to improve efficacy, reduce host toxicity and side effects, and overcome drug resistance [1]. The drug combination will present synergistic effects and may have antagonistic or additive effects [2]. In the clinical setting, synergistic effects may enable patients to be treated with a lower dose of each drug, resulting in fewer adverse side effects while still gaining the desired outcome, whereas antagonistic effects may cause patients to experience unexpected adverse side effects. Combination therapies have been explored to combat drug resistance, which cancer patients often meet under single-agent therapies [3]. Accurately predicting synergy and antagonism for drug-drug interaction (DDI) is crucial for safer and improved patient prescriptions. However, the vast number of possible drug pairs makes it difficult to screen them all experimentally. In addition, the discovery of drug combination screening by traditional experimental methods would be very challenging in terms of time, cost, and efficiency. Therefore, developing computational methods to facilitate the discovery of synergistic drug combination therapies is needed.

Chemical fingerprints can describe specific properties of drugs, including substructure, related targets, and side effects, by a series of binary digits. Of note, natural language processing methods have been utilized on SMILES (chemical language), e.g., word2vec [4] and seq2seq [5]. However, the one-dimensional (1D) sequence data could not capture the spatial structure of molecules. In other words, the constructed models cannot learn structural information directly from the input data. To deal with the lack of spatial information problem, researchers have applied graph neural networks (GNN) to obtain molecular graph representations with message passing [6]. Our input graph-like data would consist of node (atom) features and an adjacency matrix. The idea is to update each node feature vector by aggregating the message vectors passed from its neighbor nodes along the edge of the graph. In this way, we can have two-dimensional (2D) information of the drug effectively.

The advent of high-throughput sequencing enables scientists to study cancer phenotypes from cancer omics, such as genomics or transcriptomics data. The omics data are commonly used to construct cell line features for synergistic drug combination predictions as well. The cell line features play an indispensable role since the drug combination that has been validated on one cell line may not be effective on one another [7]. Jeon et al. used mutations, copy number variations, and expression of genes in cancer-related pathways to depict cell line features combined with pharmacological information to predict whether synergism or antagonism exists between two drugs [8]. Celebi et al. leveraged multi-omics data and compound properties to build a machine learning model to predict anti-cancer drug combinations [9]. Preuer et al. proposed a deep learning-based model, DeepSynergy, integrating chemical descriptors and cell line gene expression profiles to predict drug synergies [10]. AuDNNsynergy applied three autoencoders to obtain gene expression, copy number variation and mutation embeddings for individual cancer cell lines combined with physicochemical features as the input features of a deep neural network that predicts the synergy score given pairwise drug combination [11]. These methods mentioned above only consider extracting chemical properties-cell lines associations from one perspective but neglect a holistic view of interactions among features. Meanwhile, in the practical use of the model, it is sometimes difficult to obtain multiple information about the cell line except for the gene expression profiles. The protein-protein interaction (PPI) network is essential in physiological and pathological processes, including cell proliferation, differentiation, and apoptosis [12]. The potential predictability toward the drug combination by considering the drug-drug and drug-disease relationships in the PPI has been demonstrated by the network-based approach [13]. Yang et al. proposed GraphSynergy, an adapted graph convolutional network (GCN) component, to encode the higher-order topological relationship in the PPI network of protein modules targeted by a pair of drugs and the protein modules associated with a specific cancer cell line [14]. Based on GCN, PRODeepSyn integrates the PPI network with omics data to construct low-dimensional dense embedding for cell lines to make anticancer synergistic drug combination prediction [15]. However, there are still few studies considering the topological features of the drug and PPI network together in drug synergy prediction.

Attention mechanisms can improve the prediction performance and enhance the interpretability of neural network structures [16]. For the graph-structured data, the graph attention network (GAT) introduces masked self-attention layers into the node feature propagation step and multi-head attention mechanisms to stabilize the learning process [17]. Moreover, the Transformer, based solely on attention mechanisms, is beneficial to parallel computing, outperforming recurrent neural network (RNN) and convolutional neural network (CNN) [18]. Instead of chemical information-based approaches, Liu et al. used target-based representation of drug molecules inferring from drug-target associations in the PPI network to implement the TranSynergy based on attention mechanisms for synergistic prediction [19]. Wang et al. implemented AttenSyn, which exploited the attention-based pooling module to learn interactive information between drug pairs to strengthen the representations of them in synergistic drug combination prediction [20]. Utilizing the encoder of transformer to learn drug features, DeepTraSynergy is a multitask prediction model considering synergy loss, toxic loss, and drug-target interaction loss simultaneously during training synergistic drug combination prediction model [21]. Assisted by multi-layer perceptron (MLP) and GAT, DeepDDS can capture gene expression patterns and chemical substructure for identifying synergistic drug combinations toward specific cancer cell lines [22]. Based on multi-head attention mechanisms, DTSyn, a dual transformer encoder model, could capture different associations by fine-granularity transformer encoder and coarse-granularity transformer encoder for identifying novel drug combinations [23]. The AttentionDDI has been proposed to predict DDI based on a Siamese self-attention multi-model neural network that integrates multiple drug similarity measures [24]. Combining drug feature representations into four different drug fusion networks, MDF-SA-DDI predicted DDI based on the transformer self-attention mechanism [25]. By fine-tuning a pre-trained language model, DFFNDDS applied a multi-head attention mechanism and a highway network to predict synergistic drug combinations [26]. The above-mentioned studies have shown that attention mechanisms could improve not only the performance of models but also the interpretability ability in synergistic drug combination predictions.

In this study, we aim to incorporate multidimensional feature fusion methods and the attention mechanism in synergistic drug combination prediction. Here, we developed the MD-Syn (Figure 1). The MD-Syn merged both drugs and proteins' 1D and 2D representations and fed them into a fully connected neural network classifier to make a binary prediction (1: synergy, 0: antagonism). Comprehensive experiments have been designed and conducted on MD-Syn showing its feasibility for synergy drug discovery practically.

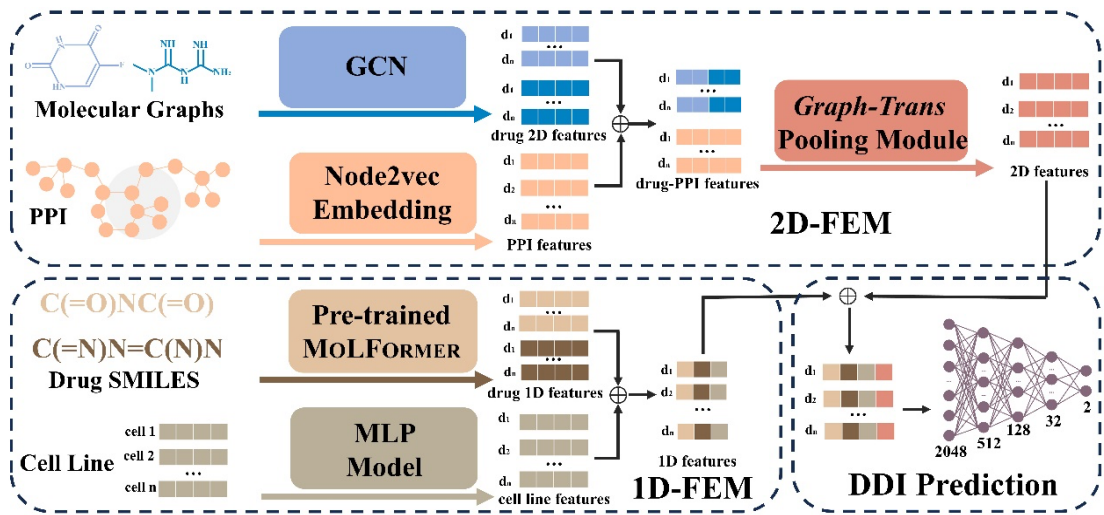

**Figure 1.** The architecture of MD-Syn. The computational framework of MD-Syn consists of a one-dimensional feature embedding module (1D-FEM), a two-dimensional feature embedding module (2D-FEM), and a fully connected neural network classifier for drug-drug interaction (DDI) prediction. For the 1D-FEM, molecular representations were obtained by fine-tuning MoLFoREMER based on the chemical SMILES; the cell line representations were compressed by a multi-layer perceptron (MLP) model. To the 2D-FEM, the molecular graph representations were learned by the graph convolutional network (GCN) module, and the protein-protein interaction network (PPI) representations were computed by the node2vec module. After combining both drug pairs and PPI network 2D representations, they would further go through a graph-trans pooling module with attention mechanisms to generate used 2D features. Consequently, by concatenating the output from 1D-FEM and 2D-FEM, we could utilize multidimensional features to train a neural network for synergistic drug combination prediction.

## Materials and methods

### Dataset

The drug-drug interaction dataset was collected from the O'Neil et al. study [27]. O'Neil dataset represents a comprehensive, unbiased, high-throughput screening of drug combinations, including 23052 drug pairs, where each pair contains two chemicals and a cancer cell line. Among the dataset, there are 39 cancer cell lines across seven different cancer types. The number of unique drugs was 38, which consists of 24 FDA-approved drugs and 14 experimental drugs [10]. The data preprocessing procedure was referred to [23].The synergy score for each drug pair was calculated using the Combenefit tool [28], which implements the well-established Loewe Additivity model to assess whether a combination exhibits synergy or antagonism. The duplicated drug pairs were averaged as a unique drug pair. Considering class balancing, 10 is a threshold to classify drug pair-cell line triplets. Triplets with synergy scores higher than 10 were positive (synergistic) pairs, indicating stronger-than-additive effects, and those less than 0 were negative (antagonistic) pairs, representing less-than-additive combined effects.

Hence, we obtained 13243 unique triplets, including 6188 positive pairs and 7055 negative pairs that covered 38 unique drugs and 31 cancer cell lines. Moreover, the gene expression profiles of cancer cell lines are from the cancer cell line encyclopedia (CCLE) [29]. As our cancer cell line features, we took the landmark genes, which can cover 82% of the whole transcriptome information in the library of integrated network-based cellular signatures (LINCS) L1000 platform [30]. Additionally, we considered two independent datasets, including Oncology Screen [27], and DrugCombDB [31] to further validate the generalization ability of the MD-Syn. All two independent datasets underwent the same data preprocessing workflow as O'Neil dataset. A total of 1919 drug synergy records were acquired for the Oncology Screen dataset, involving 21 unique drugs and 12 cell lines. To the DrugComDB dataset, there are 36626 drug combination records covering 358 drugs and 68 cell lines.

**The computational framework of MD-Syn**

In this study, we take advantage of the multidimensional feature fusion method to build a synergistic drug combination prediction model, MD-Syn. The overall architecture of MD-Syn is shown in Figure 1. The network architecture mainly contains (1) a one-dimensional feature embedding module (1D-FEM), (2) a two-dimensional feature embedding module (2D-FEM), and (3) a fully connected neural network classifier for DDI prediction. We consider drug pairs and cell line features in two different dimensional views. After concatenating the representations generated by 1D-FEM and 2D-FEM, we fed them into a fully connected neural network to make the synergistic drug combination prediction under a certain cell line. The details for each module are discussed in the following section.

**One-dimensional feature embedding module (1D-FEM) for drug pairs and cell lines**

The 1D-FEM module is in Figure 2. We first consider our input features, drug pair-cell line, in 1D view. Given the chemical SMILES for each drug, we leveraged the MoLFoReMer [32], which has been trained on over 1.1 billion molecules based on Transformer-based language model, to obtain chemical representations by fine-tuning the pre-trained model. The learned chemical representation from MoLFoReMer would be a vector with 768 dimensions. For the cell line information in the drug pair-cell line triplet, based on the 978 landmark genes for each cancer cell line, we utilized three layers of MLP to gain the compressed embedding, resulting in 256 dimensions. After that, we would merge the chemical representations of the drug pair, which learned from the MoLFoReMer, and the compressed embedding of the corresponding cancer cell line as our 1D feature.

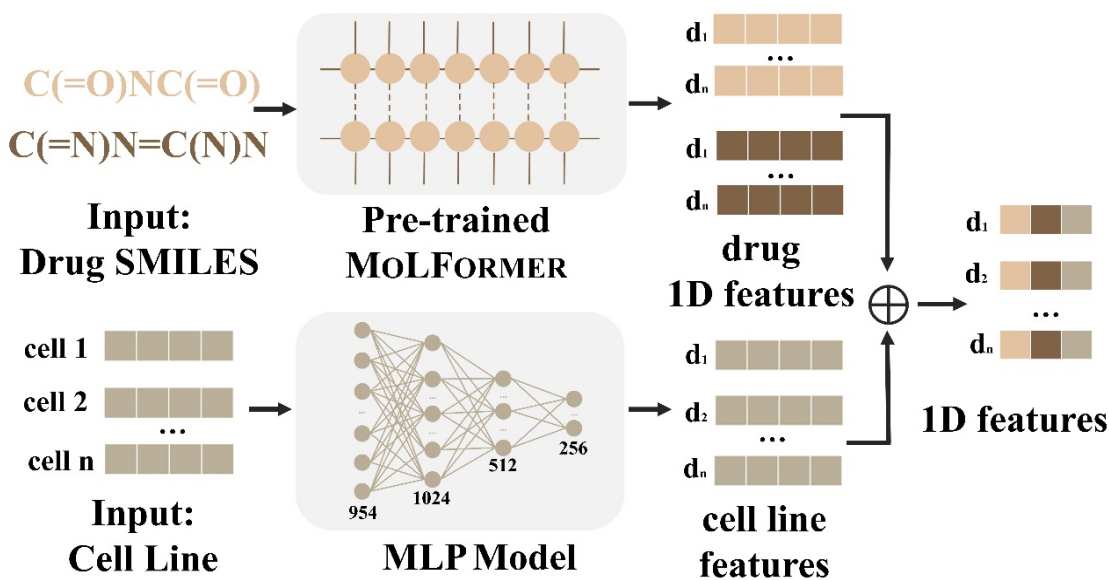

**Figure 2.** The one-dimensional feature embedding module (1D-FEM) of MD-Syn. We leverage drug pair-cell line triplets to train the MD-Syn. For one-dimensional drug features, we utilize a chemical language pre-trained model, MoLFoREMER, to obtain each drug representation based on the simplified molecular-input line entry system (SMILES) strings. The cell line information is depicted by the 978 landmark genes in the library of integrated network-based cellular signatures (LINCS) L1000 platform. They are compressed by the multi-layer perceptron (MLP). After concatenating both representations, we can have drug pair-cell line 1D features.

**Two-dimensional feature embedding module (2D-FEM) for drug pairs and cell lines**

To obtain the representation of the drug pair-cell line triplet in 2D view, we designed a 2D-FEM module, including a GCN module, a node2vec module, and a graph-trans pooling module shown in Figure 3.

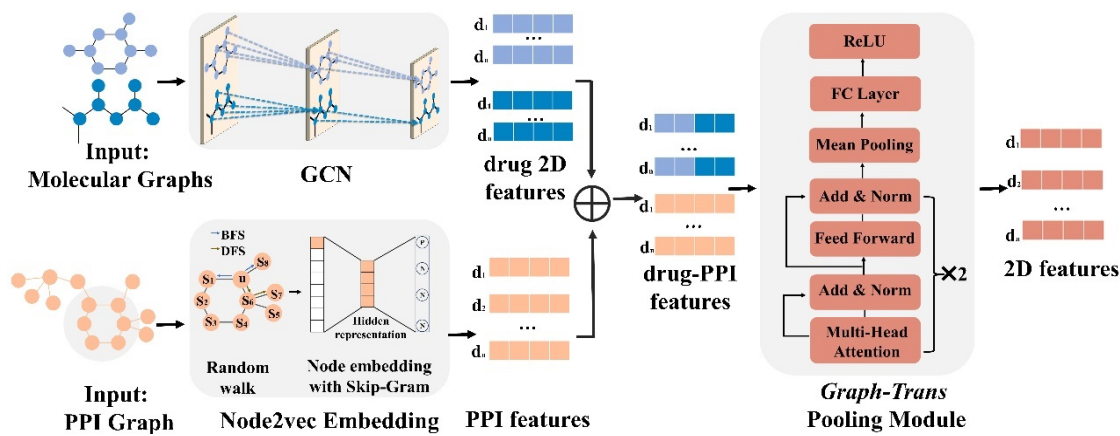

**Figure 3.** The two-dimensional feature embedding module (2D-FEM) of MD-Syn. The training dataset of MD-Syn is based on drug pair-cell line triplets. We use a graph convolutional network (GCN) to capture spatial information based on the molecular graph for each drug. Considering the PPI network, which is comprised of 978 landmark

genes, we utilize the node2vec algorithm to have each node's embedding. Given a drug pair-cell line triplet, we will concatenate the 2D feature of drug pair with PPI network embedding as the input of the graph-trans pooling module with a multi-head attention mechanism, which can generate more informative representations based on different feature aspects. We regard the output of the graph-trans pooling module as our used 2D features of the drug pair-cell line triplet.

**Graph convolutional network (GCN) in 2D-FEM for molecular graph representation learning**
Based on the RDkit [33], we could convert the SMILE format of each drug to a molecular graph. A molecular graph could be represented as $G = (V, E)$ where $V = \{v_1, v_2, ..., v_V\}$ and $E = V \times V$ denotes node and edge set, respectively. Let $X \in \mathbb{R}^{V \times c}$ be the feature matrix of all nodes, where $V$ is the number of nodes and $c$ is the dimension of node feature in a molecular graph, and $A \in \mathbb{R}^{V \times V}$ be the adjacency matrix. We would like to learn the molecular graph representation. Given a molecular graph $G$, it consists of nodes (atoms) and bonds (edges). The inputs of GCN are the node feature matrix $X \in \mathbb{R}^{V \times c}$ and the adjacency matrix $A \in \mathbb{R}^{V \times V}$. The layer-wise propagation process is shown below:

$$h^{(l+1)} = \sigma\left(D^{-\frac{1}{2}} A D^{-\frac{1}{2}} h^{(l)} W^{(l)}\right) \qquad (1)$$

where $A$ is the adjacency matrix of the molecular graph, $D_{ii} = \sum_j A_{ij}$ is a degree matrix, and $W^l$ is a layer-specific trainable weight matrix. $h^l \in \mathbb{R}^{V \times d}$ is the matrix of activation where $d$ is the compressed dimension in the $l^{th}$ layer, and $h^0 = X$. $\sigma(\cdot)$ denotes rectified linear unit (ReLU) activation function. Here, we applied a two-layer GCN followed by ReLU after each layer to capture the spatial information of the molecular graph in the GCN module.

**Node2vec in 2D-FEM for protein-protein network representation learning**
In this study, we constructed the PPI network [34] consisting of 978 landmark genes in LINCS L1000 [30]. The nodes represent proteins, and the edges indicate biological associations between proteins. To obtain the representation of each node (protein), we applied the node2vec algorithmic framework [35], which could return feature representations that maximize the likelihood of preserving network neighborhoods of nodes in a low-dimensional feature space. Given our PPI network graph, node2vec module in 2D-FEM could assist us in obtaining the feature representation in 128 feature dimensions for each node (protein).

**Graph-trans pooling with attention mechanisms in 2D-FEM for drug pair-cell line representations**

For the graph-trans pooling module in 2D-FEM, we applied two Transformer encoder layers with multi-head attention mechanisms followed by the node mean pooling strategy and a fully connected layer to generate 2D representations shown in Figure 3. The input of the graph-trans pooling module is a matrix concatenated by molecular graph representation and PPI network embeddings. An attention function maps a query and a set of key-value pairs to an output, where the query, key, and value are all from the input matrix. Here, instead of using a single attention function, we applied multi-head attention mechanisms, which projected the queries, keys, and values $h$ times with different learned query, key, and value parameter matrices. For each of the projected processes, we project our input matrix $X$ into a higher dimensional space to produce a query matrix, a key matrix, and a value matrix as below:

$$Q_i = XW_i^Q \tag{2}$$

$$K_i = XW_i^K \tag{3}$$

$$V_i = XW_i^V \tag{4}$$

where $X$ denotes the input matrix. The $W_i^Q$, $W_i^K$, and $W_i^V$ are trainable parameter matrices of the $i$-th head. $Q_i$, $K_i$, and $V_i$ stands for the query, key, and value matrices computed by the linear transformation of $X$ for the $i$-th head, respectively.

The output of the attention mechanisms for the $i$-th head is described in the following equation:

$$Attention(Q_i, K_i, V_i) = softmax\left(\frac{Q_i K_i^T}{\sqrt{d_k}}\right) V_i \tag{5}$$

where $d_k$ is the hidden dimensionality of query and key. Meanwhile, the attention score matrix is the output of the softmax activation function. In this study, we used the multi-head attention function. There are four parallel attention layers (heads) to extract associations from different aspects. Therefore, we concatenate all heads and once again project to obtain the final values as below:

$$Multihead(Q, K, V) = Concat(Head_1, Head_2, Head_3, Head_4) W^O \tag{6}$$

where $Head_i = Attention(Q_i, K_i, V_i)$ and $W^O$ is a trainable parameter matrix. It is noted the output of the multi-head attention layer would have a residual connection, which may mitigate the gradient vanishing problem, and perform layer normalization. In addition to attention sub-layers, each Transformer encoder contains a two-layer feed-forward neural network. Hence, the transformer encoder output can be depicted in the following equation:

$$TransEnc_{\text{out}} = (XW_1 + b_1) W_2 + b_2 \tag{7}$$

where $W_1$, $b_1$, $W_2$, and $b_2$ are trainable parameters. After going through two Transformer encoder layers, we did node mean pooling followed by a feed-forward neural network with the ReLu activation function to generate 2D representation, which captured the spatial information of the molecular graph and the PPI network.

**Drug-drug interaction prediction for the synergistic effect of drug combination under specified cell line**

For predicting the drug-drug interaction (DDI) with synergistic effect, we build up a MLP classifier shown in Figure 1. We are regarded the concatenation of the outputs from 1D-FEM and 2D FEM representations as, $a^0$, the input of the MLP classifier. The output of the last hidden layer of MLP classifier is $y'$ shown below:

$$y' = W_{out} \cdot a^l + b^{out} \tag{8}$$

where $W_{out}$ and $b_{out}$ are the weight matrix and bias vector, whose values would be updated by the backpropagation algorithm during the training process. The $a^l$ is the output of the previous hidden layer, which can be depicted as follows:

$$a^l = \sigma(W^l a^{l-1} + b^l) \tag{9}$$

where $W^l$ and $b^l$ are the weight matrix and bias vector of the $l$-th hidden layer. $\sigma(\cdot)$ is the ReLU activation function. Given the $i$-th drug pair-cell line triplet, to compute the probabilities of the synergistic or antagonistic effect, there is a softmax function that follows the output of the last hidden layer shown below:

$$y_i^{out} = \frac{e^{y_i'}}{\sum_{j=1}^{K} e^{y_j'}} \tag{10}$$

where $K$ denotes the number of predicted classes (synergistic or antagonistic). Furthermore, for the training process of the MLP classifier, our goal is to minimize the cross-entropy loss. It is defined as below:

$$L = -\sum_{i=1}^{N} y_i \log(y_i^{out}) \tag{11}$$

where $y_i$ is the true label of the $i$-th drug pair-cell line triplet and $N$ is the total number of training samples.

# Result

**Experimental hyperparameters setup**

MD-Syn is a computational framework that owns a significant amount of adjustable hyperparameters. It would be challenging if we exhaustively explore all hyperparameter combinations. Thus, we decided to focus on several key hyperparameters and investigated their impacts on MD-Syn performance of AUC in 5-fold cross-validation. During the hyperparameter tuning process, we found that the learning rate had the largest impact on the MD-Syn performance (Figure S1). Noted that all transformer encoder layers shared the same hyperparameters including the number of attention heads and the feed-forward hidden layer size. The searching space and the optimal hyperparameters setting are shown in Table 1.

**Table 1.** Hyperparameters of MD-Syn

| Hyperparameters | Values |
| --- | --- |
| GCN hidden units | [78,128,128]; [78,256,128]; **[78,512,128]** |
| Pooling methods | **mean**; max |
| Learning rate | 1e-2; 1e-3; 1e-4; **5e-4**; 1e-5; 1e-6 |
| Dropout rate | No dropout; 0.1; 0.2; **0.3**; 0.4; 0.5 |
| Number of attention heads | 1; **2**; 4 |
| Type of activation function | **relu**; gelu |
| Number of transformer encoder layer | 1; **2**; 3; 4 |
| Hidden size in transformer encoder | 32; **64**; 128 |
| Drug 1D feature embedding method | **MOLFOREMER**; ChemBERTa |

The bold values are the optimal parameters.

**Performance comparisons between MD-Syn and baseline methods**

To address the performance of MD-Syn, we compared MD-Syn with the state-of-the-art methods and traditional machine learning models. Among the state-of-the-art methods, the deep learning-based models contain DeepDDS [22], and DeepSynergy [10], and the Transformer-based model includes DTSyn [23]. Moreover, traditional machine learning models are random forest (RF) [36], XGBoost [37], and Adaboost [38]. The experimental results of both state-of-the-art methods and traditional machine learning models were from the same input datasets as MD-Syn. We followed the original parameter settings by referring to corresponding studies for the state-of-the-art methods, and used the default setting for the traditional machine learning models. Further, for comparing the robustness of models, all the evaluation metrics for each method were computed based on the 5-fold cross-validation shown in Table 2. For MD-Syn, the average of the area under the receiver operating characteristic curve (AUROC), area under the precision-recall curve (AUPR), accuracy (ACC), true positive rate (TPR), and F1-score (F1) are 0.919, 0.910, 0.843, 0.840, and 0.833, respectively. MD-Syn outperformed other methods with the best evaluation metrics. Specifically, compared to the transformer-based model DTSyn, it achieves a 3.9% increase in AUROC, a 4.4% increase in AUPR, and a 4.9% increase in F1, which showed that MD-Syn had superior performance in predicting synergistic drug combinations.

**Table 2.** Performance comparisons of MD-Syn and baseline methods

| Methods | AUROC | AUPR | ACC | BACC | PREC | TPR | KAPPA | F1 |
|---|---|---|---|---|---|---|---|---|
| MD-Syn | **0.919 ± 0.005** | **0.910 ± 0.006** | **0.843 ± 0.007** | **0.843 ± 0.007** | **0.827 ± 0.009** | **0.840 ± 0.014** | **0.684 ± 0.014** | **0.833 ± 0.008** |
| DeepDDS | 0.879 ± 0.021 | 0.869 ± 0.018 | 0.799 ± 0.023 | 0.797 ± 0.023 | 0.803 ± 0.019 | 0.757 ± 0.027 | 0.595 ± 0.046 | 0.779 ± 0.022 |
| DTSyn | 0.880 ± 0.004 | 0.866 ± 0.005 | 0.799 ± 0.008 | 0.798 ± 0.007 | 0.789 ± 0.011 | 0.779 ± 0.017 | 0.596 ± 0.015 | 0.784 ± 0.007 |
| DeepSynergy | 0.852 ± 0.007 | 0.841 ± 0.009 | 0.771 ± 0.008 | 0.771 ± 0.008 | 0.748 ± 0.014 | 0.779 ± 0.022 | 0.541 ± 0.016 | 0.763 ± 0.009 |
| Random Forest | 0.782 ± 0.006 | 0.822 ± 0.009 | 0.783 ± 0.006 | 0.782 ± 0.006 | 0.772 ± 0.010 | 0.760 ± 0.010 | 0.564 ± 0.013 | 0.766 ± 0.010 |
| XGBoost | 0.780 ± 0.008 | 0.822 ± 0.011 | 0.782 ± 0.007 | 0.780 ± 0.008 | 0.776 ± 0.022 | 0.752 ± 0.027 | 0.562 ± 0.015 | 0.763 ± 0.013 |
| AdaBoost | 0.761 ± 0.008 | 0.807 ± 0.007 | 0.764 ± 0.008 | 0.761 ± 0.008 | 0.763 ± 0.013 | 0.718 ± 0.021 | 0.524 ± 0.015 | 0.740 ± 0.009 |

The bold values represent the optimal parameters.

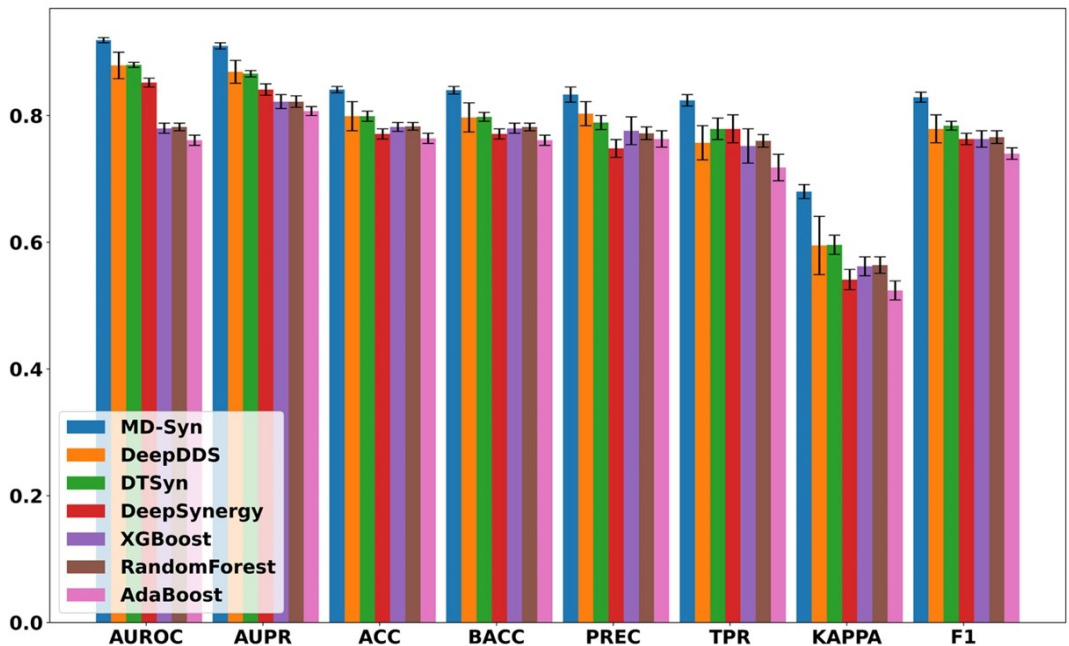

**Figure 4.** The bar plot for performance comparisons between baseline methods and traditional machine learning-based models.

**Performance evaluation based on the leave-one-out cross-validation**

In order to address the generalization ability of MD-Syn, we further performed a series of leave-one-out cross-validation experiments. Firstly, we conducted leave-drug combination-out. Based on the counts of drug pairs in our datasets, we took the former 15 drug pairs out and separated them into 5 groups as testing sets. In other words, we sequentially excluded each group and used the rest of the dataset to train MD-Syn iteratively. Moreover, other baseline methods were under the same data-splitting rule. In Table 3, MD-Syn achieved the highest average of AUROC, AUPR, and ACC of 0.865, 0.855, and 0.806, respectively, followed by DTSyn. Meanwhile, we can find that the deep learning-based and transformer-based models are significantly outperformed traditional machine learning models.

In addition, the exclusion of drug combinations during training cannot ensure that MD-Syn has not seen particular pharmaceuticals. Therefore, we would like to conduct the leave-drug-out experiment to observe the prediction ability toward the unseen drugs based on the multidimensional feature representations learned from those seen drugs. Here, we selected the top five most frequently appeared drugs in our dataset. For each fold of the training process, we only removed one out of the top five drugs. Drug pair-cell line triplets containing the specified drug would be put into the testing set and the remaining drug pair-cell line triplets would be in the training set. In Table 3, for MD-Syn, the average of AUROC, AUPR, and ACC are 0.754, 0.780, and 0.666, respectively. Meanwhile, we find that RF, XGBoost and Adaboost slightly perform better in the average of AUPR and ACC for the leave-drug-out experiment. However, MD-Syn still shows acceptable prediction performance.

To access the generalization capability on unseen cell lines, we performed leave-cell line-out experiments as well. Similarly, based on the cell line counts in our dataset, we selected the top five cell lines that appeared most frequently to form five separate testing sets. In each training fold, we selected the drug pair-cell line triplets that do not belong to the designated cell line as the training set, while the remaining triplets are used as the testing set. The average of AUROC, AUPR, and ACC are 0.804, 0.761, and 0.737, respectively. These values indicate that MD-Syn exhibits a generalization capability for cell lines that MD-Syn has not seen before.

Furthermore, we conducted experiments under more rigorous scenarios to evaluate MD-Syn. Specifically, we selected the top five tissue types that appeared most frequently and sequentially eliminated all the cell lines associated with one of them. The top five tissue types are lung, skin, intestine, ovary, and breast. During the training process, we would sequentially consider those top-five tissues once to be the testing set. Compared to baseline methods, MD-Syn has the highest average of AUROC and AUPR 0.869 and 0.852, respectively (Figure 5). Moreover, MD-Syn and other baseline methods hold a better prediction performance in intestine-correlated drug combinations within these five tissue types.

**Table 3.** Performance evaluation based on leave-drug combination-out, leave-drug-out, and leave-cell line-out.

| Methods | Leave-drug combination-out | | | Leave-drug-out | | | Leave-cell line-out | | | Leave-tissue-out | | |
|---|---|---|---|---|---|---|---|---|---|---|---|---|
| | AUROC | AUPR | ACC | AUROC | AUPR | ACC | AUROC | AUPR | ACC | AUROC | AUPR | ACC |
| **MD-Syn** | **0.865 ± 0.105** | **0.855 ± 0.098** | **0.806 ± 0.128** | **0.754 ± 0.068** | **0.780 ± 0.269** | **0.666 ± 0.158** | **0.804 ± 0.054** | **0.761 ± 0.056** | **0.737 ± 0.050** | **0.869 ± 0.016** | **0.852 ± 0.020** | **0.779 ± 0.018** |
| DeepDDS | 0.801 ± 0.099 | 0.798 ± 0.188 | 0.592 ± 0.274 | 0.735 ± 0.058 | 0.767 ± 0.239 | 0.461 ± 0.174 | 0.802 ± 0.067 | 0.744 ± 0.074 | 0.724 ± 0.044 | 0.850 ± 0.013 | 0.830 ± 0.023 | 0.767 ± 0.014 |

| | | | | | | | | | | | |
|---|---|---|---|---|---|---|---|---|---|---|---|
| DTSyn | 0.809 ± 0.066 | 0.839 ± 0.173 | 0.727 ± 0.069 | 0.733 ± 0.055 | 0.775 ± 0.216 | 0.671 ± 0.117 | 0.792 ± 0.070 | 0.749 ± 0.062 | 0.704 ± 0.031 | 0.849 ± 0.014 | 0.832 ± 0.019 | 0.750 ± 0.028 |
| DeepSynergy | 0.748 ± 0.099 | 0.792 ± 0.231 | 0.708 ± 0.095 | 0.662 ± 0.045 | 0.732 ± 0.056 | 0.462 ± 0.028 | 0.774 ± 0.075 | 0.733 ± 0.072 | 0.712 ± 0.043 | 0.842 ± 0.009 | 0.826 ± 0.017 | 0.758 ± 0.011 |
| Random Forest | 0.596 ± 0.108 | 0.835 ± 0.101 | 0.738 ± 0.06 | 0.644 ± 0.052 | **0.786 ± 0.199** | 0.621 ± 0.149 | 0.714 ± 0.056 | 0.742 ± 0.052 | 0.720 ± 0.054 | 0.780 ± 0.015 | 0.822 ± 0.022 | 0.780 ± 0.015 |
| XGBoost | 0.627 ± 0.136 | 0.827 ± 0.137 | 0.771 ± 0.048 | 0.654 ± 0.058 | **0.790 ± 0.191** | 0.644 ± 0.122 | 0.711 ± 0.054 | 0.739 ± 0.049 | 0.717 ± 0.053 | 0.768 ± 0.014 | 0.810 ± 0.020 | 0.767 ± 0.014 |
| AdaBoost | 0.612 ± 0.120 | 0.788 ± 0.195 | 0.740 ± 0.059 | 0.630 ± 0.045 | 0.775 ± 0.204 | **0.685 ± 0.110** | 0.672 ± 0.041 | 0.706 ± 0.043 | 0.677 ± 0.041 | 0.708 ± 0.014 | 0.762 ± 0.024 | 0.708 ± 0.015 |

The bold values represent the optimal parameters.

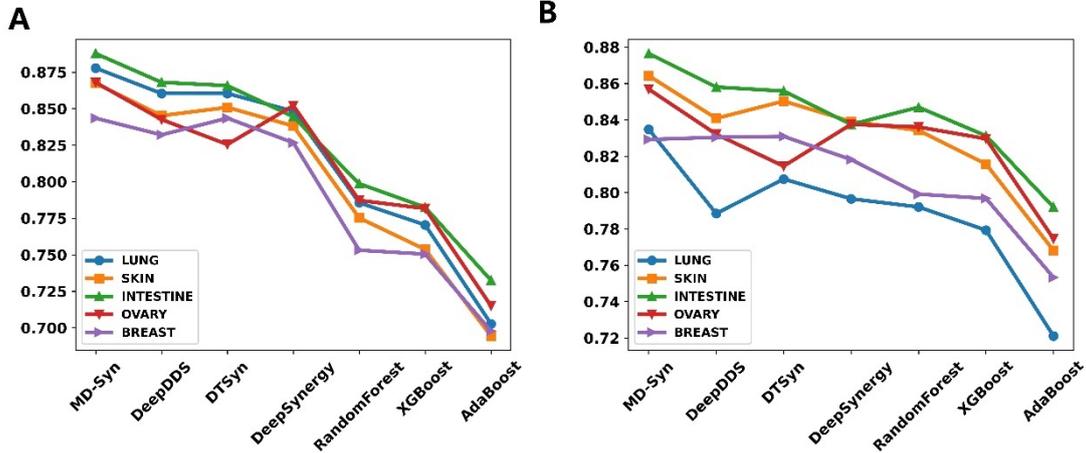

**Figure 5.** Performance results of leave-tissue-out. (A) Performance results of the average under the receiver operating characteristic curve (AUROC) for MD-Syn and baseline methods based on the top five tissue types including lung, skin, intestine, ovary, and breast. (B) Performance results of the average area under the precision-recall curve (AUPR) for MD-Syn and baseline methods based on the top five tissue types including lung, skin, intestine, ovary, and breast.

**Performance evaluation on independent datasets**

To further evaluate the generalization ability, we used O'Neil's dataset [27] to train MD-Syn and baseline methods and then leveraged the Oncology Screen [27] and DrugCombDB [31] to be the independent datasets for model validation. In Table 4, MD-Syn has the highest AUROC 0.967 and 0.625 for two independent datasets, respectively. Compared to other deep learning-based and transformer-based methods, MD-Syn demonstrated its better generalization abilities on external datasets.

**Table 4.** Performance evaluation metrics for independent datasets.

| Methods | Oncology Screen | | | DrugCombDB | | |
|---|---|---|---|---|---|---|
| | AUROC | AUPR | ACC | AUROC | AUPR | ACC |
| **MD-Syn** | **0.967** | **0.968** | **0.891** | **0.625** | **0.638** | **0.582** |
| DeepDDS | 0.900 | 0.907 | 0.804 | 0.578 | 0.582 | 0.541 |
| DTSyn | 0.921 | 0.920 | 0.841 | 0.609 | 0.635 | 0.575 |

The bold values represent the optimal parameters.

**Model ablation study**

MD-Syn takes multidimensional feature representations into account. The architecture of MD-Syn is comprised of 1D-FEM and 2D-FEM modules. The 2D-FEM module contains graph-trans pooling module introducing transformer encoder layers with multi-head self-attention mechanisms. To comprehensively investigate the contribution of each module, we conducted the ablation study for the following different combinations of modules including (1) MD-Syn with 1D-FEM only (2) MD-Syn with 2D-FEM only (3) MD-Syn with 2D-FEM without graph-trans pooling module. The corresponding results of evaluation metrics for synergistic drug combination prediction are shown in Table 5.

In MD-Syn with 1D-FEM only (MD-Syn-1D-FEM), we utilized the large-scale chimerical pre-trained model MOLFOREMER [32] to obtain 1D representation for each drug based on its SMILES. The cell line information was depicted by 978 landmark genes compressing with MLP. After concatenating those 1D representations, the later MLP classifier module would make synergistic drug combination prediction. The overall evaluation metrics are shown in Table 5. The average of AUROC, AUPR, ACC, and F1 are 0.893, 0.878, 0.809, and 0.800, respectively. Looking into MD-Syn with 2D-FEM module only (MD-Syn-2D-FEM), it focused on using 2D representations generated by GCN and node2vec algorithms based on molecular graphs and PPI network comprising 978 landmark genes, respectively. MD-Syn-2D-FEM has the average of AUROC, AUPR, ACC, and F1 0.846, 0.827, 0.764, and 0.754, respectively. Furthermore, the last variant combination is 2D-FEM without graph-trans pooling module (2D-FEM without graph-trans). The performance of 2D-FEM without graph-trans is inferior to MD-Syn-2D-FEM, indicating that multi-head self-attention mechanisms facilitate MD-Syn to capture different feature aspects resulting in better model performance. Moreover, by assisting with 2D information of input feature type, the performance of MD-Syn improves 2.6% in AUROC. The multidimensional feature representation consideration indeed leads us to achieve higher evaluation metrics in synergistic drug combination prediction. In summary, the ablation study of those three variants in 5-fold cross-validation demonstrates the importance of each component in MD-Syn and the effectiveness of graph-trans pooling in 2D-FEM module.

**Table 5.** Results of evaluation metrics for ablation study.

| Methods | AUROC | AUPR | ACC | BACC | PREC | TPR | KAPPA | F1 |
|---|---|---|---|---|---|---|---|---|
| MD-Syn | **0.919 ± 0.005** | **0.910 ± 0.006** | **0.843 ± 0.007** | **0.843 ± 0.007** | **0.827 ± 0.009** | **0.840 ± 0.014** | **0.684 ± 0.014** | **0.833 ± 0.008** |
| MD-Syn-1D-FEM | 0.893 ± 0.005 | 0.878 ± 0.006 | 0.809 ± 0.006 | 0.810 ± 0.005 | 0.783 ± 0.015 | 0.819 ± 0.021 | 0.671 ± 0.010 | 0.800 ± 0.007 |
| MD-Syn-2D-FEM | 0.846 ± 0.010 | 0.827 ± 0.017 | 0.764 ± 0.015 | 0.765 ± 0.021 | 0.734 ± 0.019 | 0.776 ± 0.073 | 0.528 ± 0.038 | 0.754 ± 0.042 |
| MD-Syn-2D-FEM-without graph-trans | 0.824 ± 0.005 | 0.808 ± 0.014 | 0.748 ± 0.010 | 0.743 ± 0.010 | 0.779 ± 0.037 | 0.649 ± 0.054 | 0.489 ± 0.020 | 0.706 ± 0.020 |

The bold values represent the optimal parameters.

**Interpretation of MD-Syn based on hidden embeddings and attention scores**

The DDI prediction module in MD-Syn is a fully connected neural network taking charge of making binary classification of synergistic or antagonistic toward the drug pair-cell line triplets. To understand whether our DDI prediction module truly learns some patterns based on the integrated multidimensional feature representations, we extracted outputs generated from 512 and 32 hidden layers. After that, we utilized a dimension reduction algorithm, UMAP [39], to project our embedding into 2D space for each drug pair-cell line triplet (Figure 6). After the training process, we found synergistic and antagonistic drug pairs were separated into two clusters distinctively. In other words, MD-Syn could identify the differences between synergistic and antagonistic effects of drug pair-cell line triplets. This phenomenon offers evidence showing that MD-Syn is capable of making synergistic drug combination predictions.

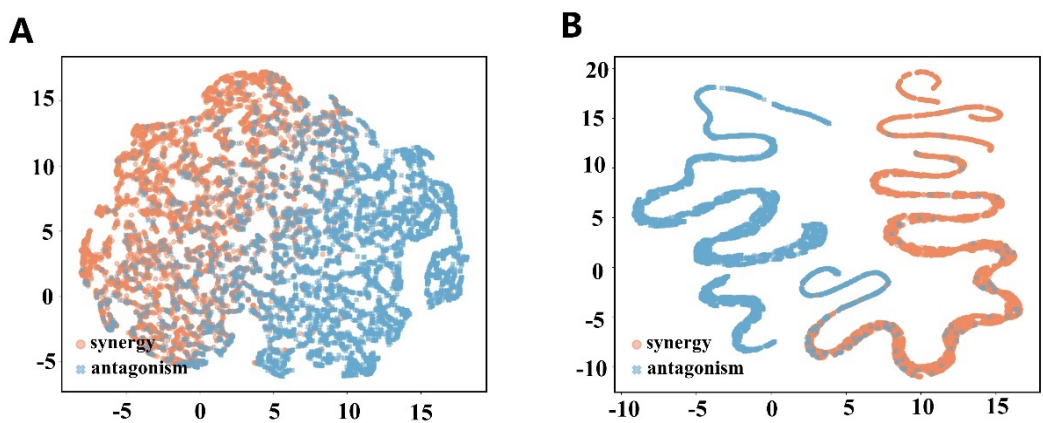

**Figure 6.** The visualization of hidden embeddings by UMAP. (A) The 512 hidden layer embedding visualized by UMAP for synergistic and antagonistic drug pair-cell line triplets. (B) The 32 hidden layer embedding visualized by UMAP for synergistic and antagonistic drug pair-cell line triplets.

The 2D-FEM contains a graph-trans pooling module built by two Transformer encoder layers with multi-head self-attention mechanisms. We analyzed the attention scores for each atom in the molecular graph and each gene in the PPI network. Here, we took two drug combinations of 5-fluorouracil (5-FU) with veliparib (ABT-888) and erlotinib, respectively, in A375 (melanoma) cell line as examples. The 5-FU, an anti-metabolite drug, is a chemotherapy drug that is widely used for cancer treatments, including colorectal, breast, pancreatic, skin, stomach, esophageal, and head and neck cancers [40]. It is particularly effective for colorectal cancer [41]. ABT-888 is a potent inhibitor of PARP activity resulting in enhancing the DNA-damaging effects of chemotherapeutic agents [42]. Erlotinib is a tyrosine kinase inhibitor of epidermal growth factor receptor (EGFR) that could block the signaling pathways involved in cancer cell proliferation, apoptosis, angiogenesis, invasions, and metastasis [43]. To our chosen examples, the drug combination of 5-FU and ABT-888 shows an antagonistic effect; the drug combination of 5-FU and erlotinib leads to a synergistic effect in the A375 cell line. After obtaining the attention score matrix based on multi-head self-attention mechanisms, we performed min-max normalization along columns. Subsequently, according to the sorting of attention scores, we identified essential elements (atoms) and highlighted them in a gradation of orange (Figures 7A-B). These elements may provide insights into the functional group interaction within the drug combination triplet. In addition, based on the attention score, we further would like to investigate how the important genes in the PPI network affect synergistic and antagonistic drug combinations. For the drug combination of 5-FU and ABT-888, 5-FU relatively pays much more attention to the *BLMH* gene, while ABT-888 is more influenced by the *ERO1A* gene (Figure 8A). The bleomycin hydrolase encoded by the *BLMH* gene is an enzyme that inactivates bleomycin, which is an essential component of chemotherapy regimens for cancer [44]. The *ERO1A* is associated with immunosuppression and dysfunction of CD8+ T cell for anti-PD-1 treatment [45]. In Figure 8B, it shows how 5-FU interacts with ABT-888 based on the attention scores for the antagonistic pair. Regarding the drug combination of 5-FU and erlotinib, the *ADGRE5* has a greater influence on 5-FU, whereas erlotinib is affected by other genes, including ADGRE1, HACD3, and PAK6 (Figure 8C). One study has demonstrated that the expression of *ADGRE5* could reflect the targeted and chemotherapeutic drug sensitivity [46]. The essential elements that the 5-FU pays attention to on erlotinib were shown on the heatmap of attention scores for the synergistic effect (Figure 8D).

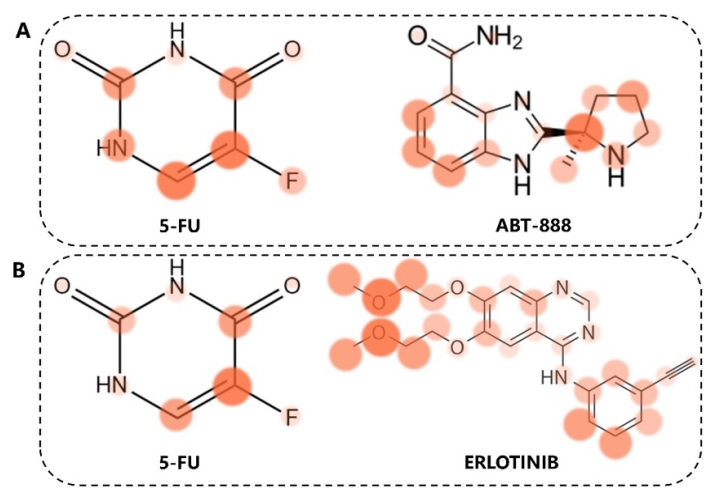

**Figure 7.** Drug combination pairs visualization results based on the attention scores. (A) The antagonistic drug combination of 5-FU and ABT-888. (B) The synergistic drug combination of 5-FU and erlotinib. The deeper the color, the more important are the elements within the drug combination pair.

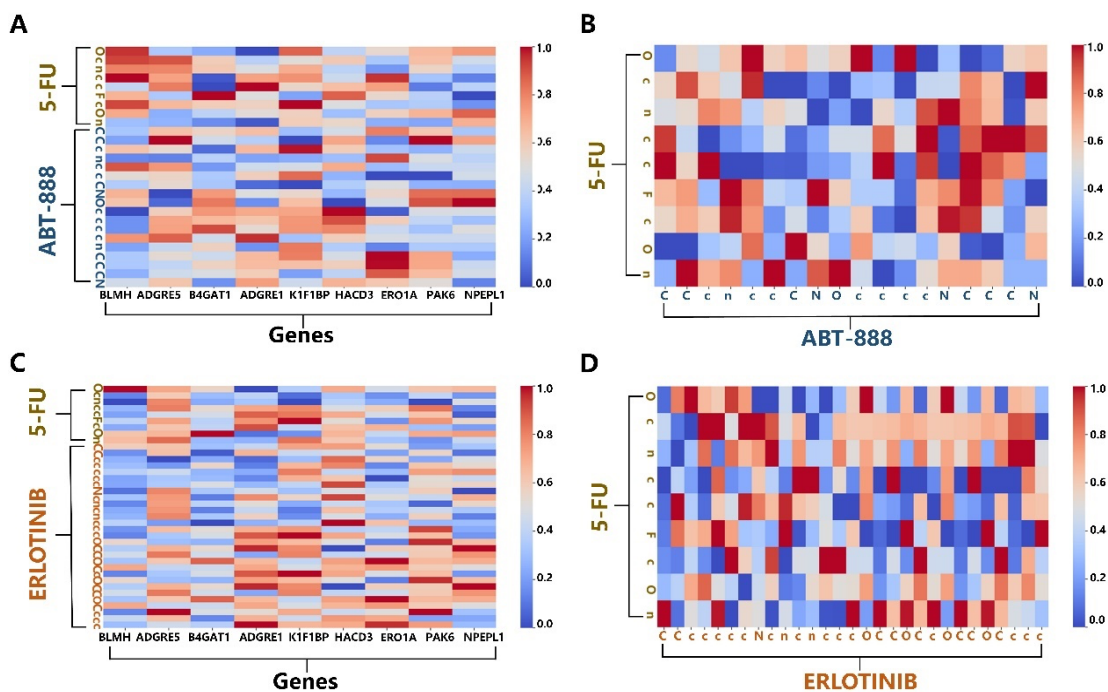

**Figure 8.** Heatmaps based on the attention scores (A) The heatmap for the drug combination of 5-FU and ABT-888 versus essential genes based on the attention scores. (B) The heatmap for 5-FU versus ABT-888 within drug combination pair based on the attention scores. (C) The heatmap for the drug combination of 5-FU and erlotinib versus essential genes based on the attention scores. (D The heatmap for 5-FU versus erlotinib within drug combination pair based on the attention scores.

In order to investigate the PPI subnetwork having more impact on identifying synergistic drug combination of 5-FU and erlotinib, we extracted genes based on the sorting of top 100 attention scores. As shown in Figure S2, we used the line width to represent the value of attention scores and the circle size to indicate the degree of node (gene).

**MDSyn: a web portal to predict the synergistic effect of drug combinations based on chemical structures and cancer cell line gene expression profiles**

To make MDSyn available to the public, we have developed a web portal based on Shiny for Python, enabling users to predict the synergistic effects of drug combinations through a web interface. The platform processes the input drugs against the model's training data using MDSyn, generating results for 1,178 drug synergy combinations: 38 (the number of O'Neil dataset unique drugs) × 31 (the number of cancer cell lines) = 1,178. Upon completion of MDSyn computation, the web interface returns combination results specific to the user-selected cell lines, providing predictions of synergistic or antagonistic effects for each drug combination. To utilize the platform, users must first obtain SMILES (Simplified Molecular Input Line Entry System) notation for their selected drugs from chemical databases such as PubChem or ChEMBL. Users then input this information and other required data (Drug Name, job title) into the web portal's information fields and select their target cell line from among 31 cancer cell lines for prediction using the MDSyn model. The prediction results are displayed in a comprehensive table showing drug combinations, cell lines, and whether their interactions are synergistic or antagonistic. For more detailed analysis, users can examine the Drug Synergy Probability Distribution interactive plot, which offers features such as adjustable displays of drug combinations through a slider bar and visualization download options. Detailed instructions for using the interactive plot can be found in the Interactive Plot Features section of the website. Users can receive comprehensive results containing all 1,178 predictions through email if a valid email address is provided. For those requiring assistance with the procedure, detailed step-by-step instructions for using MDSyn are available on the website. We believe this user-friendly MDSyn web portal will effectively accelerate wet lab drug screening processes and contribute significantly to advances in drug discovery.

## DISCUSSION

Drug combinations offer a more efficient therapeutic strategy. Dealing with numerous drug combinations, computational methods would be a faster and cheaper alternative in assisting the development of combination therapies. In this study, we introduce a novel computational framework, MD-Syn, that leverages multidimensional feature representations for drug combination prediction. Using a large-scale pre-trained model, the 1D-FEM learned the drug representations from their chemical language and obtained the embeddings for cancer cell lines from CCLE genomic profiling using MLP. Moreover, in the 2D-FEM, we learned graph representation for each drug by GCN, and had the PPI network's node embeddings by utilizing the node2vec algorithm. Our findings in the ablation study (Table 5) showed that integrating two modalities, including sequence and structural data, improved the overall performance in drug combination prediction compared to solely focusing on the feature single modality. Meanwhile, multi-head self-attention mechanisms in the graph-trans pooling module facilitate the interpretability ability of MD-Syn. By the multi-head self-attention mechanisms, MD-Syn could capture representations from different aspects of relationships among drug pair-cell line triplets.

Although MD-Syn has demonstrated outstanding performance compared to state-of-the-art methods, our proposed model has some limitations. Firstly, the training dataset that MD-Syn learned from is under Loewe score. However, there are several distinguished methods to compute expected drug combination effects from experimental data, such as combination index (CI)-isobologram equation [47], Bliss [48], ZIP score [49], and Loewe [50]. The calculated drug synergy scores would not be the same or consistent based on different quantification methods. To further improve the data quality, it is necessary to develop a new data correction method to incorporate different datasets. Secondly, the attention-based method surely provides us with a way to interpret MD-Syn. However, the atom-level coding method toward small molecules may limit the chemical interpretation. To address this limitation, merging the function-level coding method for compounds may enhance our understanding of the underlying factors that influence synergistic or antagonistic effects. Furthermore, we have found that multidimensional or multiple modalities consideration for input feature types would lead to performance improvement in synergistic drug combination prediction. Hence, integrating 3D conformation information of compounds and proteins into the graph-based model will be our future work.

In summary, MD-Syn is an innovative framework for synergistic drug combination prediction. Based on the multidimensional feature fusion method, MD-Syn significantly improves its predictive performance. Additionally, the framework offers model interpretability through multi-head attention mechanisms. By integrating multi-omics data such as transcriptomics, proteomics, and metabolomics, MDSyn could provide a more comprehensive view of drug combination effect, capturing broader biological insights and consistent performance across diverse drug-cell line combinations. This enables researchers to effectively identify potential synergistic drug combinations and advance personalized medicine for complex diseases. The framework not only advances our current understanding of drug synergy prediction but also lays a solid foundation for future developments in computational drug combination discovery and precision medicine. Beyond cancer treatment, MD-Syn's architecture demonstrates potential adaptability to other complex diseases, such as neurodegenerative disorders, where drug synergy plays a crucial role in therapeutic advancement. Furthermore, the model's flexible design shows promise for broader applications, including drug-target binding affinity prediction, extending its impact on drug discovery.

## ACKNOWLEDGMENT

The research is supported by the National Science and Technology Council (Grant number: 112-2222-E-007-001 and 113-2221-E-007-106).

## DATA AVAILABILITY

All the data used in this study were from the online database. Drug combination data were from O'Neil dataset [27]. Cancer cell line genomic data were downloaded from the DepMap portal (https://depmap.org/portal). The PPI network information was from Biogrid [34]. The source code of MD-Syn and processed data used in this research are applicable upon request.

## REFERENCES

1. Chou, T.-C., *Theoretical basis, experimental design, and computerized simulation of synergism and antagonism in drug combination studies.* Pharmacological reviews, 2006. **58**(3): p. 621-681.

2. Foucquier, J. and M. Guedj, *Analysis of drug combinations: current methodological landscape.* Pharmacology research & perspectives, 2015. **3**(3): p. e00149.
3. Jin, H., L. Wang, and R. Bernards, *Rational combinations of targeted cancer therapies: background, advances and challenges.* Nature Reviews Drug Discovery, 2023. **22**(3): p. 213-234.
4. Mikolov, T., et al., *Efficient estimation of word representations in vector space.* arXiv preprint arXiv:1301.3781, 2013.
5. Xu, Z., et al. *Seq2seq fingerprint: An unsupervised deep molecular embedding for drug discovery.* in *Proceedings of the 8th ACM international conference on bioinformatics, computational biology, and health informatics.* 2017.
6. Gilmer, J., et al. *Neural message passing for quantum chemistry.* in *International conference on machine learning.* 2017. PMLR.
7. Meng, J., et al., *Combination treatment with MEK and AKT inhibitors is more effective than each drug alone in human non-small cell lung cancer in vitro and in vivo.* PloS one, 2010. **5**(11): p. e14124.
8. Jeon, M., et al., *In silico drug combination discovery for personalized cancer therapy.* BMC systems biology, 2018. **12**(2): p. 59-67.
9. Celebi, R., et al., *In-silico prediction of synergistic anti-cancer drug combinations using multi-omics data.* Scientific Reports, 2019. **9**(1): p. 8949.
10. Preuer, K., et al., *DeepSynergy: predicting anti-cancer drug synergy with Deep Learning.* Bioinformatics, 2018. **34**(9): p. 1538-1546.
11. Zhang, T., et al., *Synergistic drug combination prediction by integrating multiomics data in deep learning models.* Translational bioinformatics for therapeutic development, 2021: p. 223-238.
12. Nero, T.L., et al., *Oncogenic protein interfaces: small molecules, big challenges.* Nature Reviews Cancer, 2014. **14**(4): p. 248-262.
13. Cheng, F., I.A. Kovács, and A.-L. Barabási, *Network-based prediction of drug combinations.* Nature communications, 2019. **10**(1): p. 1197.
14. Yang, J., et al., *GraphSynergy: a network-inspired deep learning model for anticancer drug combination prediction.* Journal of the American Medical Informatics Association, 2021. **28**(11): p. 2336-2345.
15. Wang, X., et al., *PRODeepSyn: predicting anticancer synergistic drug combinations by embedding cell lines with protein–protein interaction network.* Briefings in Bioinformatics, 2022. **23**(2): p. bbab587.
16. Wiegreffe, S. and Y. Pinter, *Attention is not not explanation.* arXiv preprint arXiv:1908.04626, 2019.
17. Veličković, P., et al., *Graph attention networks.* arXiv preprint arXiv:1710.10903, 2017.
18. Vaswani, A., et al., *Attention is all you need.* Advances in neural information processing systems, 2017. **30**.
19. Liu, Q. and L. Xie, *TranSynergy: Mechanism-driven interpretable deep neural network for the synergistic prediction and pathway deconvolution of drug combinations.* PLoS computational biology, 2021. **17**(2): p. e1008653.

20. Wang, T., R. Wang, and L. Wei, *AttenSyn: an attention-based deep graph neural network for anticancer synergistic drug combination prediction.* Journal of Chemical Information and Modeling, 2023.
21. Rafiei, F., et al., *DeepTraSynergy: drug combinations using multimodal deep learning with transformers.* Bioinformatics, 2023. **39**(8): p. btad438.
22. Wang, J., et al., *DeepDDS: deep graph neural network with attention mechanism to predict synergistic drug combinations.* Briefings in Bioinformatics, 2022. **23**(1): p. bbab390.
23. Hu, J., et al., *DTSyn: a dual-transformer-based neural network to predict synergistic drug combinations.* Briefings in Bioinformatics, 2022. **23**(5): p. bbac302.
24. Schwarz, K., et al., *AttentionDDI: Siamese attention-based deep learning method for drug–drug interaction predictions.* BMC Bioinformatics, 2021. **22**(1): p. 412.
25. Lin, S., et al., *MDF-SA-DDI: predicting drug–drug interaction events based on multi-source drug fusion, multi-source feature fusion and transformer self-attention mechanism.* Briefings in Bioinformatics, 2021. **23**(1).
26. Xu, M., et al., *DFFNDDS: prediction of synergistic drug combinations with dual feature fusion networks.* Journal of Cheminformatics, 2023. **15**(1): p. 33.
27. O'Neil, J., et al., *An unbiased oncology compound screen to identify novel combination strategies.* Molecular cancer therapeutics, 2016. **15**(6): p. 1155-1162.
28. Di Veroli, G.Y., et al., *Combenefit: an interactive platform for the analysis and visualization of drug combinations.* Bioinformatics, 2016. **32**(18): p. 2866-2868.
29. Ghandi, M., et al., *Next-generation characterization of the cancer cell line encyclopedia.* Nature, 2019. **569**(7757): p. 503-508.
30. Subramanian, A., et al., *A next generation connectivity map: L1000 platform and the first 1,000,000 profiles.* Cell, 2017. **171**(6): p. 1437-1452. e17.
31. Liu, H., et al., *DrugCombDB: a comprehensive database of drug combinations toward the discovery of combinatorial therapy.* Nucleic Acids Res, 2020. **48**(D1): p. D871-d881.
32. Ross, J., et al., *Large-scale chemical language representations capture molecular structure and properties.* Nature Machine Intelligence, 2022. **4**(12): p. 1256-1264.
33. Landrum, G., *RDKit: A software suite for cheminformatics, computational chemistry, and predictive modeling.* Greg Landrum, 2013. **8**(31.10): p. 5281.
34. Oughtred, R., et al., *The BioGRID interaction database: 2019 update.* Nucleic acids research, 2019. **47**(D1): p. D529-D541.
35. Grover, A. and J. Leskovec. *node2vec: Scalable feature learning for networks.* in *Proceedings of the 22nd ACM SIGKDD international conference on Knowledge discovery and data mining.* 2016.
36. Breiman, L., *Random forests.* Machine learning, 2001. **45**: p. 5-32.
37. Chen, T. and C. Guestrin. *Xgboost: A scalable tree boosting system.* in

*Proceedings of the 22nd acm sigkdd international conference on knowledge discovery and data mining.* 2016.

38. Freund, Y. and R.E. Schapire, *A decision-theoretic generalization of on-line learning and an application to boosting.* Journal of computer and system sciences, 1997. **55**(1): p. 119-139.
39. McInnes, L., J. Healy, and J. Melville, *Umap: Uniform manifold approximation and projection for dimension reduction.* arXiv preprint arXiv:1802.03426, 2018.
40. Longley, D.B., D.P. Harkin, and P.G. Johnston, *5-Fluorouracil: mechanisms of action and clinical strategies.* Nature Reviews Cancer, 2003. **3**(5): p. 330-338.
41. Vodenkova, S., et al., *5-fluorouracil and other fluoropyrimidines in colorectal cancer: Past, present and future.* Pharmacology & Therapeutics, 2020. **206**: p. 107447.
42. Kummar, S., et al., *Phase I study of PARP inhibitor ABT-888 in combination with topotecan in adults with refractory solid tumors and lymphomas.* Cancer Res, 2011. **71**(17): p. 5626-34.
43. Bareschino, M.A., et al., *Erlotinib in cancer treatment.* Annals of Oncology, 2007. **18**: p. vi35-vi41.
44. de Haas, E.C., et al., *Variation in bleomycin hydrolase gene is associated with reduced survival after chemotherapy for testicular germ cell cancer.* Journal of Clinical Oncology, 2008. **26**(11): p. 1817-1823.
45. Liu, L., et al., *Ablation of ERO1A induces lethal endoplasmic reticulum stress responses and immunogenic cell death to activate anti-tumor immunity.* Cell Reports Medicine, 2023. **4**(10).
46. Zhang, X., et al., *Comprehensive analysis of ADGRE5 gene in human tumors: Clinical relevance, prognostic implications, and potential for personalized immunotherapy.* Heliyon, 2024. **10**(6): p. e27459.
47. Huang, R.-y., et al., *Isobologram analysis: a comprehensive review of methodology and current research.* Frontiers in pharmacology, 2019. **10**: p. 1222.
48. Demidenko, E. and T.W. Miller, *Statistical determination of synergy based on Bliss definition of drugs independence.* PLoS One, 2019. **14**(11): p. e0224137.
49. Yadav, B., et al., *Searching for drug synergy in complex dose–response landscapes using an interaction potency model.* Computational and structural biotechnology journal, 2015. **13**: p. 504-513.
50. Lederer, S., T.M.H. Dijkstra, and T. Heskes, *Additive Dose Response Models: Explicit Formulation and the Loewe Additivity Consistency Condition.* Front Pharmacol, 2018. **9**: p. 31.